\documentclass{article}
\usepackage{arxiv}
\usepackage[utf8]{inputenc} 
\usepackage[T1]{fontenc}    
\usepackage{hyperref}       
\usepackage{url}            
\usepackage{booktabs}       
\usepackage{amsfonts}       
\usepackage{nicefrac}       
\usepackage{microtype}      
\usepackage{lipsum}
\usepackage{graphicx}
\usepackage{algorithm}
\usepackage{algorithmic}
\usepackage{multirow}

\title{Object Recognition with Human in the Loop Intelligent Frameworks}

\author{
 Orod Razeghi\\
  School of Computer Science\\
  University of Nottingham\\
  Nottingham, United Kingdom NG7 2RD\\
 \And
 Guoping Qiu\\
  School of Computer Science\\
  University of Nottingham\\
  Nottingham, United Kingdom NG7 2RD\\
}

\begin{document}
\maketitle

\begin{abstract}
Classifiers embedded within human in the loop visual object recognition frameworks commonly utilise two sources of information: one derived directly from the imagery data of an object, and the other obtained interactively from user interactions. These computer vision frameworks exploit human high-level cognitive power to tackle particularly difficult visual object recognition tasks. In this paper, we present innovative techniques to combine the two sources of information intelligently for the purpose of improving recognition accuracy. We firstly employ standard algorithms to build two classifiers for the two sources independently, and subsequently fuse the outputs from these classifiers to make a conclusive decision. The two fusion techniques proposed are: i) a modified naive Bayes algorithm that adaptively selects an individual classifier's output or combines both to produce a definite answer, and ii) a neural network based algorithm which feeds the outputs of the two classifiers to a 4-layer feedforward network to generate a final output. We present extensive experimental results on 4 challenging visual recognition tasks to illustrate that the new intelligent techniques consistently outperform traditional approaches to fusing the two sources of information.
\end{abstract}

\keywords{Information Source Fusion \and Greedy Algorithm \and Feedforward Neural Network \and Naive Bayes \and Object Recognition \and Human in the Loop}

\section{Introduction}

A persistent theme in computer vision research has been to duplicate the abilities of human visual system by electronically perceiving and understanding an image. Algorithmic methods for acquiring, processing, analysing, and understanding images from the real world to produce numerical or symbolic information in the forms of decisions are still in their infancy. For many years researchers have scrutinised every possible solution to narrow down the semantic gap between extraction of information from some visual data by computer systems and user interpretation of the same visual data in a given situation. Despite all the efforts, the simplest of many visual tasks to us humans remain a significant challenge for computers. Even in cases where the computer processing power is adequate to accomplish the task, the issue of public distrust in autonomous solutions to critical problems remains unresolved. 

The difficulties described have gradually opened a new point of view. A school of thought that considers the "human in the loop" principles as a valid solution. Having a human in the decision making loop alleviates any doubts remaining in the user's mind regarding the authenticity of a possible solution. The high-level knowledge of humans corrects mistakes made by the algorithm, whilst the technology decreases human labour in mundane tasks. Humans will be more concerned about abstract tasks, rather than tedious time consuming problems. The involvement of human in the decision making loop demands solving a number of new technical challenges. 

Amongst major human in the loop technical complications is the problem of "information source fusion". Traditionally, the fusion of low-level visual information and high-level human knowledge has been achieved in frameworks that assign equal weights to all available sources of information. Nevertheless, it is very reasonable to assume that these different origins of information are not always equally reliable or discriminative. For instance in a classification settings, outputs from a typical visual classifier may not be as accurate as the outputs from a classifier trained on user provided information. The opposite may also be valid, where user's abstract knowledge is more misleading, vague and noisy than the visual information of objects in an image. Therefore, it is very sensible to investigate either methodologies that are capable of selecting the most informative and reliable source, or algorithms that are competent enough to intelligently assign variable weights to each and every source of information available. In this paper, we mainly aim to introduce solutions to the aforementioned problem of information source fusion. Our proposed innovative algorithms are:

\begin{enumerate}
	\item A modified naive Bayes algorithm that adaptively selects an individual classifier's output or combines both to produce a definite answer.
	\item A neural network based algorithm which feeds the outputs of the two classifiers to a 4-layer feedforward network to generate a final output.
\end{enumerate}

Our proposed methods intelligently combine available sources of information in order to enhance classification performances of difficult visual recognition tasks. To illustrate the efficacy of our proposed approaches over traditional fusion techniques, we present experimental results on a variety of computer vision datasets suitable for human in the loop object recognition.


\section{Related Work}

An illustrative example in the paradigm of human in the loop is computational algorithms that find solutions to various problems interactively \cite{Qiu2010}. The proposed solution may (not) satisfy the user's expectation. If the answer is affirmative, a solution has been found. If not, the user will interact with the system to provide feedback that contains human high-level knowledge about the problem, as well as their intentions. These feedbacks will be harnessed by the algorithm as more accurate constraint conditions, or stronger priors to refine the proposed solution. This loop of human in the computational process can be iterated until a satisfactory result is achieved.

There exists a fairly limited literature on human in the loop. One of the pioneer examples may be \cite{Branson2010}, which proposes to use a Bayesian framework to combine the visual information and user answers to a set of perceptual questions for a bird species recognition system. A later work \cite{Wah2011} from the same group of researchers focuses on local part categorisation with the emphasis on users to locate different parts of an object. This model is also designed for fine-grained visual categorisation with human in the loop. By employing computer vision algorithms and analysing user responses the recognition accuracy results show an improvement over datasets of images with noisy backgrounds. It is clear from the literature that most of proposed methods struggle to intelligently fuse the two sources of low-level visual and high-level human provided information, as each component in their frameworks is estimated separately and put together subsequently with equal weights to form an answer. This kind of later fusion is the norm in the literature \cite{Razeghi2012,Razeghi2012a}, although the lack of estimating interactions between visual features and user answers is known to be an issue.

Attempts to rectify the aforementioned problem of equal weight fusion are presented in \cite{Iyengar2005,Razeghi2013,Razeghi2013a}. In a slightly different domain, authors of \cite{Iyengar2005} introduce a multimedia retrieval system that jointly models visual and textual components of a sample. As any other similar system, their human provided information is annotations from a range of users with inconsistent quality in their work, and thus the available annotations are not complete. An algorithm that evaluates the effectiveness of visual and textual components separately, and performs intelligent fusion is still desired.

We believe it is fairly worthwhile to mention that the problem of combining classifiers, which use a single source of information, has been previously studied to some extent in \cite{Kittler1998}. The authors introduce a common theoretical framework for combining classifiers by a number of simple schemes such as the product, sum, min, max, and median rule. They also compare these combination schemes to a majority voting strategy that assigns the class label based on the number of votes it receives from available classifiers. The sensitivity of these schemes to estimation errors is also investigated in their work to establish the fact that the sum rule is the most resilient combination scheme amongst the rest. In contrast to these simple rules for a single information source, our approach is to solve a supervised learning problem induced by abundance of choice in selecting the correct prediction from available classifiers that are trained separately on multiple sources of information.

In the rest of this section, we will review a number of human in the loop applications, and their potentially interactive properties in order to emphasise the importance of intelligent information source fusion in enhancing efficacy of classification tasks.


\subsection{Human in the Loop in Ecological Applications}

Torres \cite{Torres2013} presents a geo-referenced habitat image database containing over 400 high resolution ground photographs that have been manually annotated by experts. This is the first publicly available image database specifically designed for the development of multimedia analysis techniques for ecological applications. The availability of experts' annotations in this database enables human in the loop algorithms to be employed for improved categorisation of their data.

The original work from these authors presents a random forest based method for annotating an image with the habitat categories it contains. They introduce a random projection based technique for constructing their random forest classifier. Their approach is able to classify only three of the main habitat classes with a reasonable degree of confidence. Although their work has not fully examined the potential benefit of deploying a human in the loop approach, we aim to evaluate our proposed interactive methods on an extended version of their adaptable dataset.

\subsection{Human in the Loop in Medical Applications}

Existing approaches that exploit Information and Communications Technology (ICT) in dermatology, such as teledermatology (TD) and computer aided diagnosis (CAD), have had limited success in recent years. TD's total reliance on human experts viewing electronic images from a remote location to perform disease diagnosis is severely hindered by a shortage of human specialists. The core technology for CAD, computer vision, is still an evolving research subject and performances are still a long way from being practically useful. As a result, almost all research in applying CAD to dermatology was limited to diagnosing melanoma conditions and using dermatoscopic images \cite{Maglogiannis2009}. Surprisingly little research existed in applying computer vision techniques to recognising many common conditions based on ordinary photographical images.

Wide availability of mobile computing and smart phone devices have spurt extensive activities to exploit these technological advancements for dermatology applications. Carefully studying 79 dermatology-themed smart phone apps surveyed in \cite{Hamilton2012} has come to two conclusions: ubiquitous mobile computing technologies offer new opportunities and possibilities for developing new applications in dermatology to help improve patient care; however, all existing systems followed traditional TD paradigm and all have none or limited intelligent CAD capabilities.

Works in \cite{Brodley1999,Shyu1999} are possible examples of human in the loop intelligent solutions in medical image databases. \cite{Brodley1999} presents an approach to Content Based Image Retrieval (CBIR) that combines the expertise of a human, image characterisation from computer vision, and automation made possible by machine learning. Although they have an overall classification accuracy of approximately 93\%, this accuracy is not uniform across disease classes. For less populous disease classes, their accuracy can be far lower. Their solution can also benefit from a better utilisation of user feedback when retrieval results are judged unsatisfactory. \cite{Shyu1999} introduces a physician in the loop content based image retrieval system for HRCT image databases. Although the literature demonstrates a number of attempts at fabricating CBIR Medical Systems for dermatological purposes \cite{Ballerini2010a}\cite{Muller2003}, and quite a few attempts at assessing severity of specific skin diseases automatically \cite{Savolainen1998}, the lack of a reliable medical system for unskilled users, who may provide misleading information, is apparent. An ideal system should be capable of exploiting all available sources of information, whilst taking noisy data into account.

Our previous attempts in \cite{Razeghi2012,Razeghi2013} present interactive skin lesion recognition systems based on the human in the loop technology. In the papers, computer vision algorithms and models of human responses to a series of simple perceptual questions are combined together to achieve acceptable recognition rates. The proposed method in \cite{Razeghi2012} utilises a similar Bayesian framework as in \cite{Branson2010}, while \cite{Razeghi2013} exploits a random forest approach for the classification of sample images. They both introduce dermatology Q\&A banks consisting relevant perceptual questions and answers. \cite{Razeghi2012}'s database contains 10 skin conditions and 796 images, while \cite{Razeghi2013} introduces a 44 class dataset with 2309 skin images. However, none of these papers have examined the problem of information source fusion in details.

\section{Information Source Fusion}

%

\begin{figure}
	\centering
	\includegraphics[width=\linewidth]{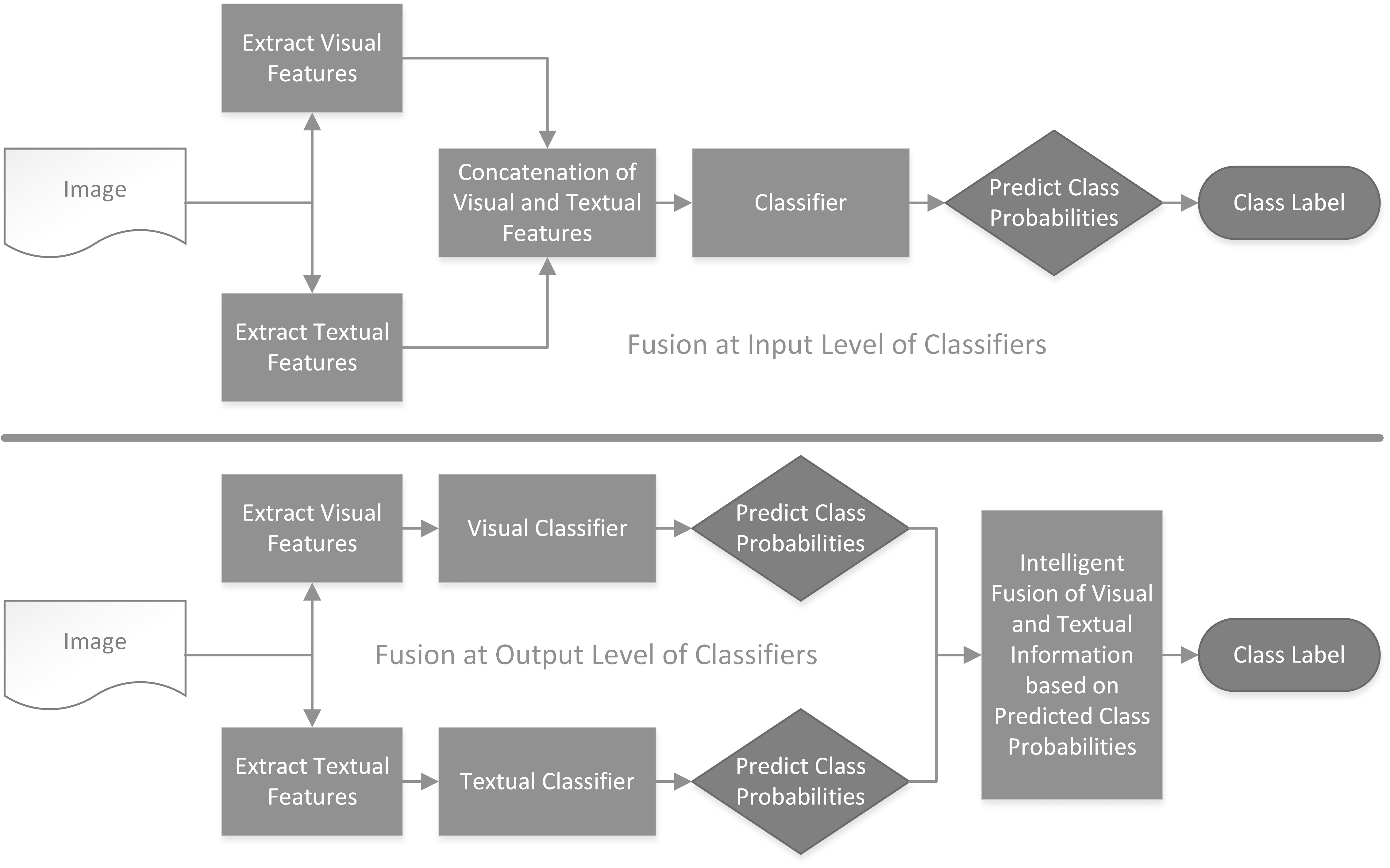}
	\caption{Fusion Frameworks at Input and Output Levels}
	\label{fig:intelFuseFrames}
\end{figure}

The problem that we aim to solve is to find the probability of an object belonging to a certain class. This is formalised in estimating a conditional probability $p(c|x,S)$ given two variables, where $c$ is class, $x$ is image information, and $S$ is any sequence of abstract information available from the human in the loop. The fusion of $x$ and $S$ is achievable both at the input or the output end of classification algorithms as illustrated in figure \ref{fig:intelFuseFrames}.

1) At input level, the fusion is performed by simply concatenating $x$ and $S$ together, and forming a universal source of information $U$. The concatenated source $U$ can be used as an input to any typical classifier. In a probabilistic settings, this is defined as:

\begin{equation}
	p(c|x,S) = p(c|x||S) = p(c|U)
\end{equation}

\noindent where $||$ is our selected notation for illustrating mathematical concatenation.


2) The fusion at output level, in contrast to the previous case, combines the output of classifiers independently trained on the two sources of information to produce an overall output in form of:

\begin{equation} \label{eq-fusion}
	p(c|x,S) = F \Big( p(c|x), p(c|S) \Big)
\end{equation}

\noindent where $F$ is the fusion function. Depending on the form of $F$, we can design a variety of fusion models. 

In spite of simplicity in implementation of fusion methods at input end of classifiers, concatenation at this level may not fully appreciate the discriminative capacity of each information source. Alternatively, we aim to propose a solution that learns separate models for each source of information available. Thus, we employ typical classifiers commonly used in the relevant literature to classify individual information sources and present two intelligent strategies to implement the fusion function $F$ of equation (\ref{eq-fusion}).

%

\section{Intelligent Information Source Fusion}

It is generally assumed that class posterior outputs from probabilistic classifiers can be considered as certainty measures, if the training and testing sets are randomly selected from the same distribution. This is a reasonable assumption for the learning problems that we are targeting to solve. Thence, this hypothesis leads us to formulation of the following two strategies.


\subsection{A Modified Naive Bayes Information Fusion Algorithm}


The information fusion in equation (\ref{eq-fusion}) becomes the classic naive Bayes classifier when the two sources of information are assumed independent. This can be formally expressed by defining the fusion function $F$ as:

\begin{equation} \label{eq-fusion-naive-Bayes}
	F(x,S) = xS \rightarrow F \Big( p(c|x), p(c|S) \Big) = p(c|x)p(c|S)
\end{equation}

\noindent where $x$ and $S$ are the image and user provided information respectively.

We have observed through experiments that for certain classes, decisions made on one information source can be more reliable than the other. It is therefore reasonable to speculate that if the probability of a class estimated from one source is too insignificant, then that source is very likely to be unreliable for predicting the class label. Based on this rationale, we present a modified algorithm of naive Bayes information fusion.

\begin{algorithm}
	\caption{Modified Naive Bayes Information Fusion $p(c|x,S)$}
	\label{algorithm1}
	\begin{algorithmic}
		\REQUIRE image information: $x$, user information: $S$, class labels: $C$
		\FORALL {Samples}
			\IF {$p(c|x) < \Theta_{x}[c]$} 
				\STATE $p(c|x,S) \propto \frac{p(c|S)}{p(c)}$
			\ELSIF {$p(c|S) < \Theta_{S}[c]$}
				\STATE $p(c|x,S) \propto \frac{p(c|x)}{p(c)}$
			\ELSE
				\STATE $p(c|x,S) \propto \frac{p(c|x)p(c|S)}{p(c)}$
			\ENDIF
		\ENDFOR
		\RETURN $\arg\max\limits_{c} p(c|x,S)$
	\end{algorithmic}
\end{algorithm}

The preceding algorithm \ref{algorithm1} is very straightforward. For each class $c$, we estimate a threshold $\theta_{x}[c]$ for image's visual information source, and a separate threshold $\theta_{S}[c]$ for user's abstract source of information. If the probability of a class estimated from one source is smaller than its threshold, then only the probability estimated based on the other source is used to predict the class. When the probabilities of a class estimated from both sources are greater than their respective thresholds, the original naive Bayes is utilised. If the estimated probability is smaller than both thresholds, it does not matter which classifier is employed. We will illustrate in the experimental section that for certain applications this modification can significantly improve accuracy over the classic naive Bayes classifier, which is indeed a special case of our intelligent fusion algorithm.

%

The optimal thresholds for each class and every source of information are estimated by a grid search approach that exhaustively examines a range of possible values. The selected threshold for each class is a value that leads to the finest classification performance over the training dataset of samples with known class labels $T = \{ (I_i,C_i) : i\in[n] \}$. This is achieved by minimising empirical risk:

\begin{equation}
	L(\Theta;T) = \frac{1}{n}\sum^n_{i=1} l(C_i,f(I_i;\Theta))
\end{equation}

where $\Theta$ is set of thresholds to be learned, and $l$ measure of error between groundtruth $C_i$ and predicted $f(I_i;\Theta)$ labels. Threshold values in practice filter out uncertain predictions from deployed classifiers in our fusion framework. Algorithm \ref{algorithm2} summarises our method in selecting suitable thresholds.

\begin{algorithm} [t]
	\caption{A Grid Search Approach for Optimal Threshold Selection}
	\label{algorithm2}
	\begin{algorithmic}
		\REQUIRE class label: $c$, matrix of posterior probabilities: $P$
		\STATE Step 0: Generate a discrete set of possible thresholds: $\Theta = \{0,0.1,0.2,...,1\}$
		\STATE Step 1: Create an empty set to store scores of each threshold: $Scores = \emptyset$
		\WHILE {there exist unexamined $\theta \in \Theta$}
			\STATE Step 2: Create an empty set to store predicted labels: $R = \emptyset$
			\FORALL {samples with true label $c$}
				\STATE Step 3: Find vector of posterior probabilities: $V = P_{j,:}$
				\STATE Step 4: Find probability of the most probable class: $p = \arg\max(V)$
				\IF {$p > \theta$}
					\STATE Accept label: $R = R \cup label$
				\ENDIF
			\ENDFOR
			\STATE Step 5: Calculate $F_1$ measure obtained by threshold $\theta$
			\STATE Step 6: Store calculated measure: $Scores = Scores \cup measure$
		\ENDWHILE
		\RETURN $\arg\max\limits_\theta(Scores)$
	\end{algorithmic}
\end{algorithm}

\subsection{Neural Network Fusion Algorithm}

The method discussed previously is a greedy approach. It follows the problem solving heuristic of making the locally optimal choice in selecting suitable weights for every predicted class label with the hope of finding a global optimum. The weights are in essence the calculated thresholds for outputs from classifiers. However, greedy algorithms usually fail to find the globally optimal solution. Our proposed greedy approach does not operate holistically on all class labels. It examines each class label at a time, and hence can make commitments to certain choices too early, which prevents it from finding the best overall solution afterwards. It may even produce the unique worst possible solution.

In an attempt to find the global optimum, we propose a pattern recognition solution that trains a supervised neural network to produce desired outputs in response to sample inputs. More specifically, we intend to deploy a feedforward backpropagation network \cite{Werbos1974}. Our selected choice of network training function is a scaled conjugate gradient backpropagation approach \cite{Moller1993} that updates weight and bias values according to the scaled conjugate gradient method.

\begin{figure} [htbp]
	\centering
	\includegraphics[width=\linewidth]{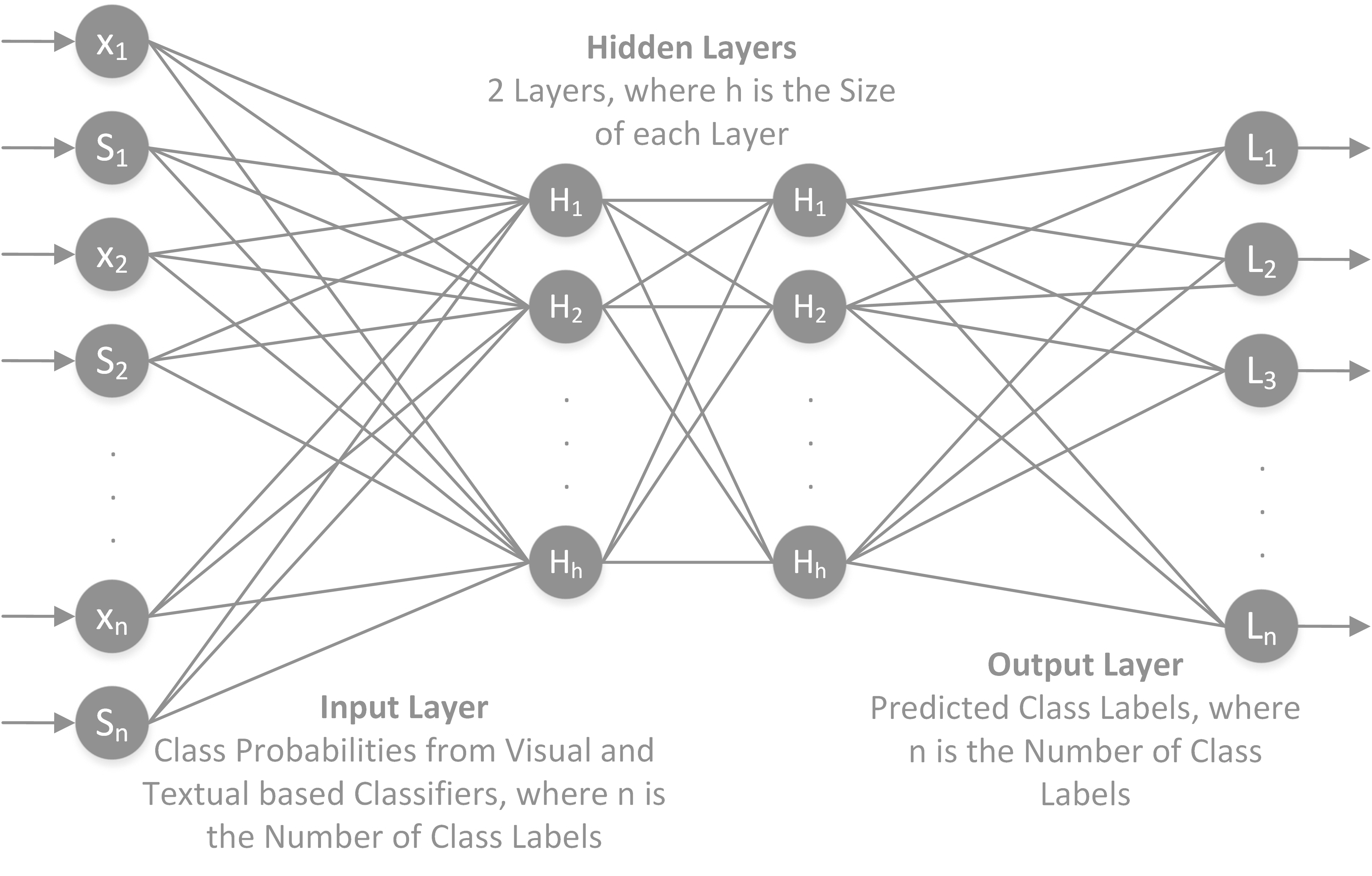}
	\caption{Neural Network Layout}
	\label{fig:net}
\end{figure}

The architecture of the neural network we aim to use has 4 layers. This is depicted in figure \ref{fig:net}. The input layer has $2n$ input units. $i_k$, where $k = \{1,2,...,n\}$, is the predicted probability of class $k$ based on the image's visual source of information using a standard classifier. Some of these classifiers are described in section \ref{sec:cls} of this paper. Similarly, $S_k$, where $k = \{1,2,...,n\}$, is the predicted probability of class $k$ based on the user provided textual source of information. We intend to employ two hidden layers in our implementation and the number of units can be determined experimentally. The output layer has $n$ units, each corresponding to one of the class labels. In preparing the desired output for an input training sample, we set the corresponding unit's desired output to $1$ and the rest to $0$. For instance:

\begin{equation}
	L_1=0, L_2=0, ..., L_{k-1}=0, L_k=1, L_{k+1}=0, ..., L_n=0
\end{equation}
 
\noindent is the desired output corresponding to a training sample belonging to class $k$. Once the network is trained, the final decision about the class label is made based on the following:

\begin{equation} \label{eq-nn-decision}
	c^* = \arg\max\limits_{k} L_k
\end{equation}


We further need to calculate a network performance that leads to good classification. Thus, we suggest to minimise a cross-entropy term given targets, outputs, performance weights, and optional parameters. Our selected entropy term is therefore defined as:

\begin{equation}
	H(C,\hat{C}) = - \sum_i C(i) \log\hat{C}(i)
\end{equation}

\noindent where $H(C,\hat{C})$ is the computed cross-entropy of true and predicted class labels, which heavily penalises outputs that are extremely inaccurate, with very little penalty for fairly correct classifications.

\section{Classifiers in Fusion Frameworks} \label{sec:cls}

The classification of available information from images and users involved in human in the loop vision applications is usually performed by a number of commonplace techniques. Fusion at input level for instance can be carried out by an SVM solution or an ensemble approach like random forests that concatenates visual and textual descriptors. At output level, it is plausible to apply a combination of similar or different classifiers on available sources separately. Table \ref{tab:settings} summarises these common settings in human in the loop frameworks. Please note that RF and RNB stand for Random Forest and Random Naive Bayes respectively in this table, and $||$ is the selected notation for illustrating mathematical concatenation of descriptors.

\begin{table} [htbp]
	\centering
	\caption{Classifiers Settings in Human in the Loop Frameworks}
	\begin{tabular}{c|c|c|c}
		\hline
		Feature      & Visual  & Textual      & Visual $||$ Textual\\
		\hline\hline
		Input Level  & N/A     & N/A          & SVM, RF\\
		Output Level & SVM, RF & SVM, RF, RNB & N/A\\
		\hline
	\end{tabular}
	\label{tab:settings}
\end{table}

In the next section, we introduce our proposed generative classifier suitable for harnessing user abstract knowledge, and our selected discriminate method for classification of low-level visual information in human in the loop applications. In the experiment section of this paper, we will compare the result of these classifiers to their common alternatives.

\subsection{A Generative Model for User High-Level Information} \label{sec:rnb}

We build an innovative random naive Bayes model to estimate posterior probability $p(c|S)$. This allows us to classify user's high-level information effectively. This model learns the class-conditional density $p(S|c)$ for each value of $c$. It is basically a generative model that estimates user answers $S$ for each possible class label $c$.

\textbf{High-level Information Presentation:} As in our previous works \cite{Razeghi2012,Razeghi2012a,Razeghi2013,Razeghi2013a}, we collect high-level information about images in the form of answers to perceptual questions. These answers can be regarded as presence of tags in each image. The importance of these tags become apparent when visual features fail to capture the complexity presents in visually similar images.

Suppose there are $T$ possible tags in our problem. Let $t\in\{1,...,T\}$ be an array of indices to those $T$ tags, and let $S = \{s_1,...,s_{t-1},s_t\}$ be a set of answers from user about presence of such tags in an image. Then an image can be represented as a vector of tags presence.

To deal with user reasoning that is approximate rather than exact, we quantify presence of tags in an image by a certainty value that describes user confidence in their response. This is in contrast to the traditional binary approach, where tags' random variables take on only true or false values. Our answered tags random variable $s_t$ have a discrete truth value that ranges in an interval between $0$ and $1$, corresponding to their chance of presence in an image. These certainty values $k\in\{1,...,K\}$ allow the model to assign more weights to more confident answers. Any positive answer has a probability value above $0.5$, and any negative one is below $0.5$. We set analogous terms for these certainty values like "probably" as a middle value between "definitely" and $0.5$, and "guessing" as a middle value between "probably" and $0.5$. Table \ref{tab:cert} shows these certainty values, where $K = 6$. Although other definitions of these numerical values are possible, we have not considered this as a focus of current studies.

\begin{table}[htbp]
	\centering
	\caption{User Answers Certainties} \label{tab:cert}
	\begin{tabular}{c|c|c|c}
		\hline
		Answer   & Guessing & Probably & Definitely\\
		\hline\hline
		Positive & 0.625    & 0.75     & 1\\
		Negative & 0.375    & 0.25     & 0\\
		\hline
	\end{tabular}
\end{table}

\textbf{Modelling User Answers:} Our proposed generative model for estimating user provided high-level information needs the class conditional density $p(S|c)$ to be specified. We make the assumption that questions are answered by user independently given the class, and any randomness in their response is not image dependant:

\begin{equation}
	p(S|c) = \prod^{T}_t p(s_t|c)
\end{equation}

We estimate $p(s_t|c)$ separately for each value of $c$, thus we only solve $C$ separate density estimation problems. An expedient strategy to avoid the problem of exponential explosion is to naively assume that the parameters of such conditional distribution are independent. Since $s_t\in\{1,...,K\}$, the multinomial class-conditional density for each $p(s_t|c)$ is defined as:

\begin{equation}
	p(s_t|c,\theta_c) = \prod^K_{k=1} \theta^{I(s_t=k)}_{ck}
\end{equation}

Similarly, we fit a multivariate multinomial distribution to our discrete training set of vectors:

\begin{equation}
	p(S|c,\theta) = \prod^T_{t=1} \prod^K_{k=1} \theta^{I(s_t=k)}_{tck}
\end{equation}

\noindent where $\theta_{tck} = p(s_t=k|c)$ is the probability of observing the $t^{th}$ tag being $k$ given that the class label is $c$.

\textbf{Ensemble of Random Naive Bayes Classifiers:} Frequently an ensemble of models performs superiorly in contrast to any individual model. To take advantage of such increased stability, we propose to employ an ensemble averaging process, where we train a group of random naive Bayes Classifiers. Each of these classifiers has a low bias and high variance. The combination of these classifiers usually results in a new network with low bias and low variance.

In an ensemble learning method, injection of randomisation leads to decorrelation between the individual classifiers, and improved generalisation. We introduce randomisation in our ensemble model by two mainly used means of: i) random input selection, and ii) random feature selection. These procedures of randomisation injection help us achieve higher robustness with respect to presence of noisy data in user responses.

Firstly, given our collected training set $S$ of size $n$, a bootstrap aggregating technique is used to generate $m$ new training sets $S_i$, each with the same size $n$, by sampling from $S$ uniformly and with replacement. By sampling with replacement, some observations may be repeated in each $S_i$. Every set $S_i$ is expected to have the usual fraction $\sim 63.2\%$ of the unique examples of $S$, the rest being duplicates. The $B$ random naive Bayes models are subsequently fitted using the above $m$ bootstrap samples and combined together by a voting scheme for final classification.

Secondly, we randomly select $F$ features out of user responses' feature pool $S$ for each random naive Bayes classifier. The probability distribution $p(s_f|c)$ of each feature is therefore modelled for each class $c$. The probability of a sample $s$ belonging to the class $c$ can eventually be defined as:

\begin{equation}
	p(s|c) \sim \sum^B_{b=1} \prod^F_{f=1} p^b(s_f|c)
\end{equation}

\noindent where we combine $B$ randomly trained naive Bayes classifiers, each using a subset of available features $F <= |S|$.

\subsection{A Discriminative Classifier for Image Low-Level Information} \label{sec:rf}

In our proposed implementation, the visual model of the fusion equation (\ref{eq-fusion}) is designed to directly learn a function that computes the class posterior $p(c|x)$. This is therefore defined as a discriminative model that discriminates between different classes given the visual feature input.

Image representation plays an important role in the quality of any visual classification solution. It is believed that a careful combination of visual descriptors may improve performance of the classification algorithm but this is not the main focus of our current work. Instead, we aim to use a selection of well-known visual words with specific parametrisation to form visual feature vectors suitable for our classifiers.

Our main selected discriminative model of visual information is based on a bootstrap aggregating ensemble algorithm that follows the standard method in \cite{Breiman2001} to train random trees, and classify test samples. 
The widely adopted information gain criteria, calculated based on class labels of the training images, is used as the score function to select a good split:

\begin{equation}
	Score(split) = \triangle E=-\frac{|I_{l}|}{|I_{n}|}E(I_{l})-\frac{|I_{r}|}{|I_{n}|}E(I_{r})
\end{equation}

\noindent where $E(I)$ is the Shannon entropy of class labels distributions in the set of samples $I$. $I_l$ and $I_r$ represent the training images contained in node $n$'s left and right child nodes respectively. $I_n$ is the set of training sample in node $n$. Leaf nodes store a normalised probability distribution of the occurrence of all possible classes in the dataset.

\section{Experiments and Results}

We now illustrate the effectiveness of our proposed intelligent fusion techniques suitable for incorporating human abstract knowledge in the decision making loop of visual object recognition tasks. We have tested our solution on 4 datasets appropriate for evaluating human in the loop applications. There are two applications of fine-grained visual classification, and two examples from medical settings.

The selected classifiers in this section are based on two discriminative methods for visual and textual features: i) an RBF kernel Support Vector Machine, and ii) an ensemble of 1000 bagged decision trees that we described previously in \ref{sec:rf}. We also have two generative approaches suitable only for textual information from users: i) method of \cite{Branson2010}, and ii) our introduced 1000 random naive Bayes solution in \ref{sec:rnb}. All results presented are based on a 5-time repeated random sub-sampling cross validation method. It may be necessary to clarify that the fusion accuracies at input level presented for random forest and SVM methods are the results of concatenating visual and textual feature vectors.

\subsection{Caltech-UCSD Birds 200 Dataset}

CUB-200 \cite{Welinder2010a} is a dataset of 6033 images over 200 bird species, such as Myrtle Warblers, Pomarine Jaegars, and Black-footed Albatrosses. These classes cannot usually be identified by users with no prior expertise. Different bird species are nearly visually identical in many cases of this dataset. A set of 25 visual questions that encompass 288 binary answers are available in the dataset. The presence of binary attributes in an image is determined by users answering the visual questions. There are 15 training images per class in the dataset, and the remaining 3033 images are used for evaluation purposes. Bird training and testing images are roughly cropped by a bounding box.

\textbf{Visual Results:} The computer vision algorithm in \cite{Branson2010} is based on Andrea Vedaldi's publicly available source code \cite{Vedaldi2009}, which combines vector-quantised geometric blur and colour/grey SIFT features using spatial pyramids, multiple kernel learning, and per-class 1-vs-all SVMs. The authors also add features based on full image colour histograms and vector-quantised colour histograms. They use a validation set to tune parameters for the visual classification $p(c|x)$. 
For comparative purposes, we also test this dataset using our proposed discriminative random forest solution with the same visual features.

The main advantage of employing computer vision on this dataset is to reduce human labour by minimising the number of questions user has to answer, or in other words the number of tags needed to improve the quality of classification predictions. Computer vision is more effective at reducing the average amount of time than reducing the time spent on the most challenging images. It is clear from the results in table \ref{tab:birds} that random forest outperforms the SVM algorithm on this dataset.

\textbf{Textual Results:} A deterministic user with precise responses is assumed to achieve perfect classification accuracy on this dataset within the first few rounds of answering questions. However, this assumption is not realistic, since subjective answers by user are common and unavoidable. Stochastic user responses increase the number of questions necessary to achieve a certain accuracy level. It is important to note that some images in this dataset can never be classified correctly without computer vision, and solely by utilising user answers.

We represent the mean classification accuracy results when merely user responses are incorporated without any computer vision involved in the process in this section. The learned model of \cite{Branson2010} based on multinomial distribution results in a mean accuracy of 66\% due to its ability to tolerate a reasonable degree of error in user answers. We also include the results from a number of different methods capable of estimating class conditional $p(S|c)$ to clearly illustrate the power of our random naive Bayes solution. The performances of an SVM baseline solution in addition to the discriminative random forest method in \cite{Razeghi2013,Razeghi2013a} are also included for better comparison in table \ref{tab:birds}.

\begin{table}[htbp]
	\centering
	\caption{Mean Accuracy of Classification Algorithms on CUB-200} \label{tab:birds}
	\begin{tabular}{c|l l}
		\hline
		Information Source             & Classification Technique       & Mean Accuracy\\
		\hline\hline
		\multirow{2}{*}{Visual Based}  & SVM                            & 19\%\\\cline{2-3}
		                               & Random Forest                  & \textbf{20.51\%}\\\hline\hline
		\multirow{4}{*}{Textual Based} & SVM                            & 61.92\%\\\cline{2-3}
		                               & Random Forest                  & 66.43\%\\\cline{2-3}
		                               & Naive Bayes \cite{Branson2010} & 66\%\\\cline{2-3}
		                               & Random Naive Bayes             & \textbf{68.89\%}\\\hline
	\end{tabular}
\end{table}

\textbf{Combination Results:} The fusion of information sources at input level with feature concatenation and output level with assigned equal weights show no significant difference. However, it is clear from table \ref{tab:birds-comb} that our intelligent source selection methods outperform the conventional fusion techniques. Our neural network method of intelligent source selection based on predictions from a random forest visual classifier and a random naive Bayes textual classification technique yields the best performance at 68.89\%. This is in contrast to baseline results from authors in \cite{Branson2010}, who report an average accuracy of 66\% based on a SVM visual classifier and a naive Bayes textual classification method.

It is also worth to mention that user responses drive up the accuracy of computer vision algorithms. Not only vision improves overall performance but also there are some cases that can not be correctly classified without computer vision, even after asking all possible questions from user. The main advantage of the visual questions paradigm is that contextual sources of information can easily be incorporated in the system. For instance in this dataset, information such as behaviour and habitat can be utilised as additional questions to help with better identification of different species. It is clear that our fusion techniques improve the overall accuracy of this dataset more effectively than conventional methods.

\begin{table}[htbp]
	\centering
	\caption{Mean Accuracy of Classification Algorithms on CUB-200} \label{tab:birds-comb}
	\begin{tabular}{c|c|l|l|l}
		\hline
		Fusion                        & Classifier & \multicolumn{3}{l}{Concatenation}\\
		\hline\hline
		\multirow{2}{*}{Input Level}  & SVM        & \multicolumn{3}{l}{63.32\%}\\\cline{2-5}
		                              & RF         & \multicolumn{3}{l}{66.32\%}\\\hline\hline
		Fusion                        & Classifier & Equal Weight & Greedy Alg.      & Neural Net.\\
		\hline\hline
		\multirow{3}{*}{Output Level} & SVM+SVM    & 63.38\%      & \textbf{64\%}    & \textbf{64.18\%}\\\cline{2-5}
		                              & RF+RF      & 66.38\%      & \textbf{67.3\%}  & \textbf{68.7\%}\\\cline{2-5}
		                              & RF+RNB     & 66.4\%       & \textbf{68.83\%} & \textbf{68.89\%}\\\hline
	\end{tabular}
\end{table}

\subsection{Ground Photograph Habitat Dataset}

The extended version of Ground Photograph Habitat database \cite{Torres2013} consists of 1086 ground images with 4203 annotated polygons. There are 27 distinct habitats present in the dataset. The numbers of instances (shown inside the brackets) from each habitat are as follows: Woodland broad-leaved (399), Woodland mixed (242), Scrub dense (295), Scrub scattered (21), Acid grassland semi-improved (149), Neutral grassland unimproved (125), Neutral grassland semi-improved (386), Improved grassland (296), Marshy grassland (62), Poor semi-improved grassland (6), Bracken continuous (55), Bracken scattered (16), Tall ruderal (52), Dry dwarf shrub heath acid (40), Dry dwarf shrub heath basic (8), Dry heath acid grassland mosaic (88), Fern (1), Standing water (18), Cultivated arable (63), Cultivated ephemeral short perennial (1), Hedge and trees species rich (110), Hedge and trees species poor (226), Fence (231), Wall (11), Dry ditch (15), Sky (1042), Other (non-habitats) (245).

Mostly habitats from classes: Woodland and scrub, Grassland and marsh, Tall herb and fern, Heathland, and Miscellaneous that includes the boundary habitats are present in this database. All photographs are manually ground-truthed by an expert in Phase 1 classification. To incorporate the high-level information provided by human in the loop, an expanded version of this dataset introduces a set of 17 binary perceptual questions that summarises the perceptual information of image as seen by the human. The questions are listed in table \ref{tab:habit-qa}. These questions were answered by inexpert users.

\begin{table}[htbp]
	\centering
	\caption{Ground Photograph Habitat Dataset Questions} \label{tab:habit-qa}
	\begin{tabular}{l|l}
		\hline
		\multicolumn{2}{c}{Tags used in the Questions}\\
		\hline\hline
		1. Heath    & 10. Trees with leaves\\
		2. Water    & 11. Trees without leaves\\
		3. Sky      & 12. Trees with mixed leaves\\
		4. Bushes   & 13. Arable land or crops\\
		5. Wall     & 14. Grass with flowers or non-uniform grass\\
		6. Fence    & 15. Uniform grass\\
		7. Reed     & 16. Bracken or fern\\
		8. Herbs    & 17. Other\\
		9. Boundary & \\
		\hline
	\end{tabular}
\end{table}

The habitat dataset is a multilabel classification problem. It is a variant of the multiclass classification problem, where multiple target labels must be assigned to each instance. Formally, multilabel learning can be phrased as the problem of finding a model that maps inputs $x$ to vectors $y$, rather than scalar outputs as in the ordinary classification problem. To make the dataset compatible with the scalar outputs of our multiclass classification algorithms, we exploit a transformation method that maps each combination of labels present in the dataset to a unique new label. This translates to 347 unique class labels. We split the dataset to 657 training, and 429 testing images. Evaluation metrics for multilabel classification are inherently different from those used in multiclass classification, due to the inherent differences of the classification problem. In our paper, we use the following metrics for the habitat dataset:

\begin{itemize}
	\item Hamming Loss (Relaxed Metric): is the percentage of the wrong labels to the total number of labels. This is a loss function, so the optimal value is zero. $1 - loss$ equals to the accuracy.
	\item Exact Match (Strict Metric): is the most strict metric, indicating the percentage of samples that have all their labels classified correctly.
\end{itemize}

\textbf{Visual Results:} We build 8976-dimensional visual feature vectors to represent the visual information of the habitat dataset. The visual features used are: Coloured Pattern Appearance Model (CPAM) \cite{Qiu2002}, Geometric Blur (GB) \cite{Vedaldi2009}, Global Image Descriptor (GIST) \cite{Oliva2001}, Pyramid Histogram of Oriented Gradients (PHOG) and its variations \cite{Vedaldi2009}, Scale-invariant Feature Transform (SIFT) and its variations, Pyramid Histogram of Visual Words (PHOW) and its variations \cite{Vedaldi2008}, and Self-similarity Feature (SSIM) \cite{Vedaldi2009}.

The low-level visual features in this dataset particularly struggle to distinguish between semi-improved, and unimproved grassland classes of this dataset. These classes are even subjective for human surveyors. Additionally, broad-leaved trees can be part of both the Broad-leaved Woodland habitat, which is composed of broad-leaved trees, and the Mixed Woodland habitat, which is itself composed of broad-leaved trees and coniferous trees. This similarity in classes explains the reason why the low-level features may struggle to classify these habitats. It is evident that our proposed random forest classifier outperforms the alternative baseline SVM significantly.

\textbf{Textual Results:} The answer to questions in table \ref{tab:habit-qa} can be used to build user response pairs suitable for the Bayesian framework of \cite{Branson2010}. Each user response $s$ contains an answer $a$, and a confidence value $r$ that deals with user's uncertainty in answering the questions. The answers to the questions can also be used to build textual feature vectors of 17 dimension suitable for our approach. The results of these modelling methods can be found in table \ref{tab:habit}. It is again clear that our random naive Bayes method surpasses other possible solutions.

\begin{table}[htbp]
	\centering
	\caption{Mean Accuracy of Classification Algorithms on Ground Photograph Habitat Dataset} \label{tab:habit}
	\begin{tabular}{c|l l}
		\hline
		Information Source             & Classification Technique      & Mean Accuracy\\
		\hline\hline
		\multirow{2}{*}{Visual Based}  & SVM                           & 38.91\% (3.03\%)\\\cline{2-3}
		                               & Random Forest                 & \textbf{56.6\% (16.78\%)}\\\hline\hline
		\multirow{4}{*}{Textual Based} & SVM                           & 44.22\% (5.94\%)\\\cline{2-3}
		                               & Random Forest                 & 52.81\% (11.65\%)\\\cline{2-3}
		                               & Naive Bayes \cite{Branson2010}& 45.66\% (6.06\%)\\\cline{2-3}
	                                   & Random Naive Bayes            & \textbf{58.72\% (17.94\%)}\\\hline
	\end{tabular}
\end{table}

\textbf{Combination Results:} Our proposed intelligent source selection methods prove to be very effective on this dataset. The neural network approach based on predictions from random forest visual and random naive Bayes textual classifiers achieves an intriguing mean accuracy of 68.25\% and 35.19\% for relaxed and strict metrics respectively. These results highlight the fact that there is an impressive 15.47\% improvement over results from the same classifiers joined in an equal weight framework for the relaxed metric. This is also true for the strict metric, where a 23.25\% enhancement is evident. The baseline results based on the proposed algorithm in \cite{Branson2010} yield an average accuracy of 51.21\% for relaxed and 11.65\% for strict metrics.

Table \ref{tab:habit-comb} summarises the accuracies of different fusion methods. The results of both evaluation metrics we described previously is presented. As it is clear, the combination of low-level visual features with high-level knowledge of users increases the average accuracy of the algorithms. This is true with both metrics. The addition of human high-level knowledge drives up the mean accuracy of the algorithm. It is obvious that our intelligent fusion techniques are outperforming other frameworks in every aspect of the evaluation.

\begin{table}[htbp]
	\centering
	\caption{Mean Accuracy of Classification Algorithms on Ground Photograph Habitat Dataset. Representing both Relaxed and (Strict) Metrics} \label{tab:habit-comb}
	\begin{tabular}{c|c|l|l|l}
		\hline
		Fusion                                      & Classifier           & \multicolumn{3}{l}{Concatenation}\\
		\hline\hline
		\multirow{2}{*}{Input Level}                & SVM                  & \multicolumn{3}{l}{50.32\% (10.78\%)}\\\cline{2-5}
		                                            & RF                   & \multicolumn{3}{l}{57.22\% (17.94\%)}\\\hline\hline
		Fusion                                      & Classifier           & Equal Weight                   & Greedy Alg.                            & Neural Net.\\
		\hline\hline
		\multirow{3}{*}{\shortstack{Output\\Level}} & \scriptsize{SVM+SVM} & \scriptsize{51.14\% (11.2\%)}  & \scriptsize{\textbf{53.3\% (12.01\%)}} & \scriptsize{\textbf{59.88\% (21.63\%)}}\\\cline{2-5}
		                                            & \scriptsize{RF+RF}   & \scriptsize{57.22\% (17.94\%)} & \scriptsize{\textbf{59.7\% (21.47\%)}} & \scriptsize{\textbf{65.22\% (30.78\%)}}\\\cline{2-5}
		                                            & \scriptsize{RF+RNB}  & \scriptsize{52.78\% (11.94\%)} & \scriptsize{\textbf{61.6\% (22.37\%)}} & \scriptsize{\textbf{68.25\% (35.19\%)}}\\\hline
	\end{tabular}
\end{table}

\subsection{DERM2309 Skin Condition Dataset}

This dataset \cite{Razeghi2014} contains images of skin conditions from 44 different diseases. In the original release of this dataset, there are 20 training images per class and the rest are used for testing. There are 880 training and 1429 testing images, totalling 2309 images in the dataset. The lesions are manually segmented using a bounding box. Extracted visual features are included in the public release of the dataset. Skin lesion images range from different types of Eczema to various cancerous conditions, such as Superficial Spreading Melanoma. Rare conditions like Bullous Pemphigoid, and more common diseases like Psoriasis are amongst the condition classes of this dataset. The dataset contains 10 questions and 37 possible binary answers that summarise the patient's skin lesion characteristics. Workers from a crowd-sourcing tool have provided answers to these perceptual questions. The answers are useful for building classification models. Table \ref{tab:derm} illustrates classification accuracies using different methods and sources of information.

\textbf{Visual Results:} The SVM solution has a mean classification accuracy of 13.37\%. The random forest technique results in an average accuracy of 15.76\%. These classifiers are both fed with the same type of visual features available from the public release of this dataset.

\textbf{Textual Results:} The SVM classifier using textual features results in an accuracy of 14.77\%. A random forest trained with the textual features has an average accuracy of 16.58\%. The learned model of \cite{Branson2010} based on multinomial distribution results in a mean accuracy of 18.4\% on this dataset. Our proposed random naive Bayes method has an average accuracy of 21.02\%, a better performance than the SVM, and random forest approach. This reiterates the effectiveness of our proposed solution to modelling user high-level information in this dataset.

\begin{table}[htbp]
	\centering
	\caption{Mean Accuracy of Classification Algorithms on DERM2309} \label{tab:derm}
	\begin{tabular}{c|l l}
		\hline
		Information Source             & Classification Technique      & Mean Accuracy\\
		\hline\hline
		\multirow{2}{*}{Visual Based}  & SVM                           & 13.37\%\\\cline{2-3}
		                               & Random Forest                 & \textbf{15.76\%}\\\hline\hline
		\multirow{4}{*}{Textual Based} & SVM                           & 14.77\%\\\cline{2-3}
		                               & Random Forest                 & 16.58\%\\\cline{2-3}
		                               & Naive Bayes \cite{Branson2010}& 18.4\%\\\cline{2-3}
		                               & Random Naive Bayes            & \textbf{21.02\%}\\\hline
	\end{tabular}
\end{table}

\textbf{Combination Results:} The most effective fusion solution for this dataset is once again proved to be the neural network approach based on random forest visual and random naive Bayes textual classification techniques. It is quite interesting to note that in table \ref{tab:derm-comb} a simple concatenation of visual and textual features and utilising a SVM algorithm yields a classification accuracy of merely 16.03\%. However, the deployment of neural network approach based on the same visual and textual features results in a mean accuracy of 34.81\%. This is an improvement of 18.78\%. Baseline's \cite{Branson2010} combination accuracy result of this dataset saturates at 22.39\% using a visual SVM and a textual naive Bayes classification technique.

Once visual and textual features are combined at input level using the random forest technique, the classification accuracy rises to 25.12\%. These improvements show the usefulness of high-level knowledge in shape of answers from users. Our most effective version of the greedy solution outperforms the Bayesian baseline combination in \cite{Branson2010} by approximately 8\%. Our neural network approach performs better than other fusion techniques with a 34.81\% mean accuracy. These results also illustrate the importance of source selection, and our proposed intelligent fusion methods.

\begin{table}[htbp]
	\centering
	\caption{Mean Accuracy of Classification Algorithms on DERM2309 Dataset} \label{tab:derm-comb}
	\begin{tabular}{c|c|l|l|l}
		\hline
		Fusion                        & Classifier & \multicolumn{3}{l}{Concatenation}\\
		\hline\hline
		\multirow{2}{*}{Input Level}  & SVM        & \multicolumn{3}{l}{16.03\%}\\\cline{2-5}
		                              & RF         & \multicolumn{3}{l}{25.12\%}\\\hline\hline
		Fusion                        & Classifier & Equal Weight & Greedy Alg.      & Neural Net.\\
		\hline\hline
		\multirow{3}{*}{Output Level} & SVM+SVM    & 18.7\%       & \textbf{20.98\%} & \textbf{23.9\%}\\\cline{2-5}
		                              & RF+RF      & 25.58\%      & \textbf{28.7\%}  & \textbf{31.74\%}\\\cline{2-5}
		                              & RF+RNB     & 26.08\%      & \textbf{30.95\%} & \textbf{34.81\%}\\\hline
	\end{tabular}
\end{table}

\subsection{MIAS Mammographic Dataset}

As another potential human in the loop medical application, we test our algorithms on the MIAS database released by the Mammographic Image Analysis Society \cite{JSuckling1994}. The owners of this dataset have clipped or padded every image to 1024 pixels by 1024 pixels. Their public release of dataset contains the films, and appropriate high-level information as follows: character of background tissue (Fatty, Fatty Glandular, Dense Glandular), class of abnormality (Calcification, Well-defined/circumscribed masses, Spiculated masses, Other, ill-defined masses, Architectural distortion, Asymmetry, Normal), and severity of abnormality (Benign, Malignant). The dataset has 90 training, 232 testing, and a total of 322 images. The images in the dataset can be grouped into: benign, malignant, and normal classes. A set of 2 possible questions and 10 binary answers are available in this dataset that can be used to incorporate the high-level knowledge of human in the loop. Table \ref{tab:mias-qa} lists these binary questions.

\begin{table}[htbp]
	\centering
	\caption{MIAS Dataset Questions} \label{tab:mias-qa}
	\begin{tabular}{l|l}
		\hline
		\multicolumn{2}{c}{Tags used in the Questions}\\
		\hline\hline
		Background Tissue           & Type of Abnormality\\\hline
		1. Calcification            &  8. Fatty\\
		2. Well-defined masses      &  9. Fatty-glandular\\
		3. Needle-like masses       & 10. Dense-glandular\\
		4. ill-defined masses       & \\
		5. Architectural distortion & \\
		6. Asymmetry                & \\
		7. Normal                   & \\
		\hline
	\end{tabular}
\end{table}

\textbf{Visual Results:} We build visual feature vectors of 5756 dimension to represent the visual information of images in this dataset. In addition to the SIFT and PHOG features, we add Grey-Level Co-occurrence Matrix (GLCM) \cite{Haralick1973}, Local Binary Patterns (LBP) \cite{Vedaldi2008}, Local Phase Quantisation (LPQ) \cite{Ojansivu2008}, and Canny Edge Detector \cite{Canny1986} descriptors to the selection.

In this dataset, the low-level visual features struggle mostly between the benign, and malignant classes. It seems that it is very hard to distinguish between these two classes using only visual features. The random forest classifier is outperforming the SVM baseline result on this dataset using the same visual features, as seen in table \ref{tab:mias}.

\textbf{Textual Results:} The answer to the binary questions can be used to build user response pairs suitable for the Bayesian method described in \cite{Branson2010}. Each user response $s$ contains an answer $a$, and a confidence value $r$. As before, to utilise the random forest classifier, the questions are exploited to build a textual feature vectors of 10 dimension. Random forest performs almost as effective as the naive Bayes model in \cite{Branson2010} using merely textual features. Our proposed random naive Bayes method produces comparable mean accuracies to other accurate approaches on this dataset.

\begin{table}[htbp]
	\centering
	\caption{Mean Accuracy of Classification Algorithms on MIAS Dataset} \label{tab:mias}
	\begin{tabular}{c|l l}
		\hline
		Information Source             & Classification Technique      & Mean Accuracy\\
		\hline\hline
		\multirow{2}{*}{Visual Based}  & SVM                           & 14.65\%\\\cline{2-3}
                               		   & Random Forest                 & \textbf{28.44\%}\\\hline\hline
		\multirow{4}{*}{Textual Based} & SVM                           & 88.79\%\\\cline{2-3}
		                               & Random Forest                 & 88.36\%\\\cline{2-3}
		                               & Naive Bayes \cite{Branson2010}& 89.65\%\\\cline{2-3}
                                	   & Random Naive Bayes            & 89.65\%\\\hline
	\end{tabular}
\end{table}

\textbf{Combination Results:} Table \ref{tab:mias-comb} summarises the mean accuracies of various techniques using different sources of information. The combination of low-level visual features with high-level knowledge of human in the loop leads to an average accuracy of 89.65\% using the Bayesian method \cite{Branson2010} of equal weight fusion. Due to the discriminative nature of questions in this dataset, the mean accuracy of the framework using textual features is very high. However, the Bayesian fusion method fails to exploit the information contained in the visual features to improve the combination accuracy. The fusion accuracy is only as precise as the tags' results on this dataset using the Bayesian or alternative techniques. Our introduced greedy algorithm of source selection can improve the performance slightly but our neural network approach exploits the visual information more efficiently, and enhances the fusion accuracy to 94.81\% from the baseline result of 89.65\% produced by the Bayesian framework.

At input level, the random forest framework concatenates the visual and textual feature vectors. The addition of human high-level knowledge drives up the mean accuracy of the algorithm from 28.44\% based on visual descriptors to a staggering 90.94\%. It is clear from the results that visual features alone achieve very low recognition rates, reiterating the challenging nature of these visual tasks. Nevertheless, human in the loop knowledge can boost recognition rates to more acceptable levels. Our proposed fusion techniques produce enhanced results to most accurate solutions on this dataset.

\begin{table}[htbp]
	\centering
	\caption{Mean Accuracy of Classification Algorithms on MIAS Dataset} \label{tab:mias-comb}
	\begin{tabular}{c|c|l|l|l}
		\hline
		Fusion                        & Classifier & \multicolumn{3}{l}{Concatenation}\\
		\hline\hline
		\multirow{2}{*}{Input Level}  & SVM        & \multicolumn{3}{l}{88.65\%}\\\cline{2-5}
		                              & RF         & \multicolumn{3}{l}{90.94\%}\\\hline\hline
		Fusion                        & Classifier & Equal Weight & Greedy Alg.      & Neural Net.\\
		\hline\hline
		\multirow{3}{*}{Output Level} & SVM+SVM    & 88.45\%      & \textbf{89.25\%} & \textbf{89.25\%}\\\cline{2-5}
		                              & RF+RF      & 89.23\%      & \textbf{90.4\%}  & \textbf{91.23\%}\\\cline{2-5}
		                              & RF+RNB     & 90.9\%       & 90.08\%          & \textbf{94.81\%}\\\hline
	\end{tabular}
\end{table}

\section{Conclusion}

In this paper, we introduced novel intelligent methodologies for selecting the most effective source of information available in human in the loop fusion frameworks. It is very interesting to note that our proposed neural network approach always produces superior, or at minimum comparable results to the greedy method of selecting information sources. It is also important to reiterate the fact that both our intelligent information fusion techniques improve mean classification accuracies of currently common literature methods such as: feature concatenation at input level, and considering equal weights for separate classifiers at output level.

We believe that our intelligent method of source selection plays a deciding role in effective incorporation of human in the loop knowledge that is truly necessary in solving difficult tasks of object recognition. Our proposed approaches effectively select the most reliable source of information from available classifiers and fuse them to produce more reliable predictions. Moreover, our introduced random naive Bayes solution to modelling user answers is a novel and efficient method in the relevant human in the loop literature. The experimental results illustrate the effectiveness of our solutions on a variety of application domains.

\bibliographystyle{unsrt}  
\bibliography{manuscript}  

\end{document}